\documentclass[letterpaper]{CVIS}
\usepackage{amsmath,bm}
\usepackage{graphicx}
\usepackage{dcolumn}
\usepackage{subcaption,soul}
\usepackage{caption}
\usepackage[numbers,sort&compress,sectionbib]{natbib}
\usepackage{tabularx} 
\usepackage{multirow} 
\usepackage{makecell} 
\usepackage{url}
\usepackage{mleftright}
\usepackage{multirow}
\usepackage{titlesec}
\usepackage{amssymb}
\usepackage{float}
\usepackage{booktabs} 
\usepackage[redeflists]{IEEEtrantools}
\usepackage[pagebackref=false,breaklinks=true,colorlinks=true,linkcolor = blue, urlcolor = black,citecolor=black,bookmarks=true]{hyperref}

\title{A Unified Model Selection Technique for Spectral Clustering Based Motion Segmentation}

\begin{document}
\bstctlcite{IEEEexample:BSTcontrol}
\sloppy
\author{
Yuxiang Huang$^{1}$ \quad 
John Zelek$^{1}$ \\
$^1$Vision and Image Processing Lab, System Design Engineering, University of Waterloo\\
\texttt{\{yuxiang.huang, jzelek\}@uwaterloo.ca}\\
}

\maketitle
\begin{abstract} 
Motion segmentation is a fundamental problem in computer vision and is crucial in various applications such as robotics, autonomous driving and action recognition. Recently, spectral clustering based methods have shown impressive results on motion segmentation in dynamic environments. These methods perform spectral clustering on motion affinity matrices to cluster objects or point trajectories in the scene into different motion groups. However, existing methods often need the number of motions present in the scene to be known, which significantly reduces their practicality. In this paper, we propose a unified model selection technique to automatically infer the number of motion groups for spectral clustering based motion segmentation methods by combining different existing model selection techniques together. We evaluate our method on the KT3DMoSeg dataset and achieve competitve results comparing to the baseline where the number of clusters is given as ground truth information. 
\end{abstract}

\section{Introduction}

The objective of motion segmentation is to divide a video frame into regions segmented by common motions. Currently, motion segmentation is still a challenging problem when a moving camera is present, due to unknown camera motion. One popular technique to solve the motion segmentation problem in such scenario is to perform spectral clustering on motion affinity matrices constructed with motion models \cite{brox_object_2010, vidal_subspace_2011, li_perspective_2013, ochs_segmentation_2014, xu_motion_2018, jiang_what_2021, xi_multi-motion_2022, huang_motion_2023, lin_multi-motion_2023}. These methods typically take manually corrected point trajectories as input and build custom motion affinity matrices using one or more types of motion cues such as geometric models, spatio-temporal similarities or optical flow. Recently, spectral clustering based methods have shown remarkable results in segmenting motions in challenging dynamic environment containing significant motion degeneracy and complex scene structures \cite{xu_motion_2018, jiang_what_2021, xi_multi-motion_2022, lin_multi-motion_2023, huang_motion_2023}, largely thanks to its ability of synergetically fusing multiple types of motion cues together. However, all of these methods cannot automatically infer the number of motions present in the scene (i.e., model selection) and rely on user input for such information.  \cite{brox_object_2010, vidal_subspace_2011, li_perspective_2013, ochs_segmentation_2014} do propose model selection techniques, but those techniques are specifically suited for their respective methods, which do not perform well in complex dynamic scenes. To address this issue, we propose a general unified model selection technique by combining the strengths of multiple existing criteria, to automate the model selection process for the current spectral clustering based motion segmentation methods relying on either single or multiple types of motion affinities.

\section{Methodology}
We first briefly introduce the motion segmentation method being used as a foundation and baseline for our model selection technique, then discuss the proposed model selection technique in detail.

\subsection{Motion Segmentation}
\begin{figure*} [thbp]  
    \centering
    \includegraphics[width=\textwidth]{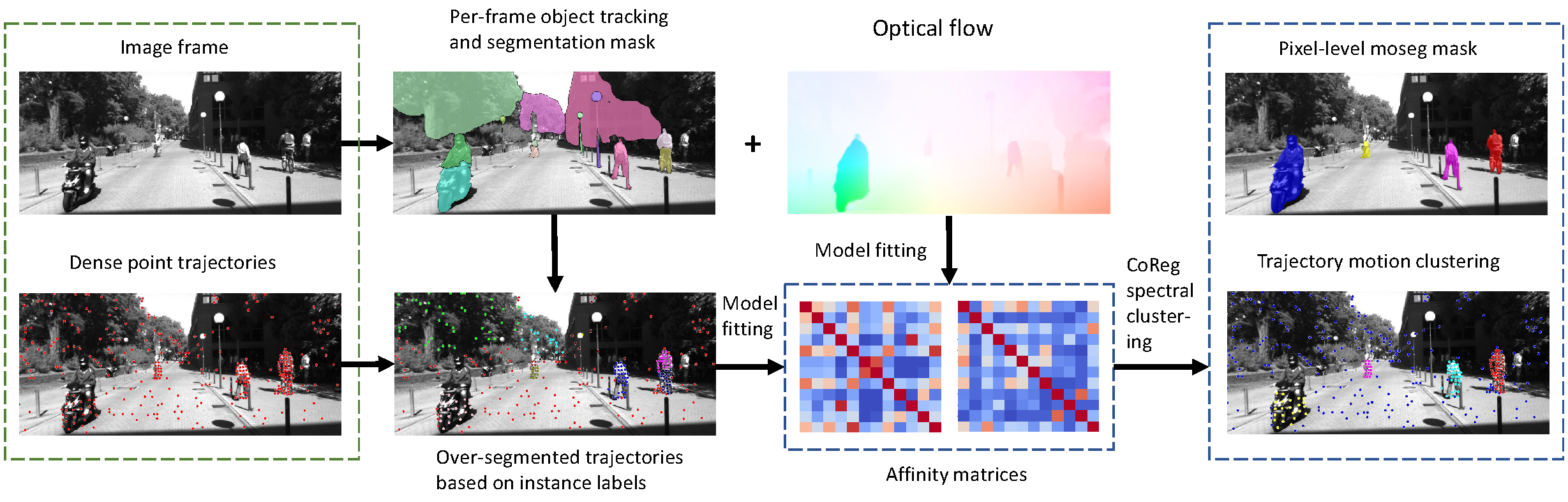} 
    \caption{Motion Segmentation Pipeline. Given a sequence of video frames, 1) generate an object proposal for every frame, 2) obtain object-specific point trajectories and optical flow as two types of motion cues, 3) construct two motion affinity matrices using pair-wise object motion affinities, 4) perform co-regularized spectral clustering on the two motion affinity matrices to obtain the final segmentation}
    \label{fig: Motion Segmentation Pipeline}
\end{figure*}

We use our previously proposed motion segmentation method \cite{huang_motion_2023} as the baseline. \cite{huang_motion_2023} performs motion segmentation by clustering different objects into different motion groups according to their pairwise motion similarities. More specifically, it first generates an object proposal for every frame of the video sequence denoting all common objects present in the scene, using a combination of off-the-shelf object recognizer, detector, segmentor and tracker. After all the potential objects in the video are segmented and tracked, object-specific point trajectories and optical flow mask for each labeled object in the video are generated as motion cues. From these two types of motion cues, two robust affinity matrices are constructed to encode the pairwise object motion affinities throughout the whole video using epipolar geometry and the optical flow based parametric motion model. Finally, co-regularized multi-view spectral clustering is used to fuse the two affinity matrices and obtain the final clustering. Figure \ref{fig: Motion Segmentation Pipeline} shows a diagram of this motion segmentation pipeline. This method achieves state-of-the-art results on the challenging KT3DMoSeg dataset by fusing multiple motion models together using multi-view spectral clustering, similar to other recent methods. Therefore, it is an ideal baseline to evaluate our model selection method.

\subsection{Model Selection}
We propose a general unified model selection method by combining four widely used model selection methods, i.e., the silhouette score \cite{rousseeuw_silhouettes_1987}, eigengap heuristic \cite{von_luxburg_tutorial_2007}, Davies-Bouldin index \cite{davies_cluster_1979} and Calinski-Harabasz index \cite{calinski_dendrite_1974}, to obtain an improved accuracy in determining the number of motion groups in the scene. We choose to use these four methods since they are all widely used criteria to evaluate the quality of clustering as well as to determine the optimal number of clusters. Given a motion affinity matrix, we first compute a confidence score for each criterion on a range of possible number of motions that may be present in the scene, we then compute the average of all four confidence scores corresponding for every possible number of motions, and select the one with the the highest confidence as the number of clusters to perform spectral clustering. We briefly introduce these four model selection criteria and further discuss our proposed method in the following sections. 

\subsubsection{Silhouette Score}
The silhouette score measures how closely related each sample is to other samples in the same cluster comparing samples in other clusters. A higher silhouette score indicates higher similarity among samples within each cluster and lower similarity among samples in different clusters, hence better clustering quality. The mean Silhouette score for the clustering can be written as follows:
\begin{equation}
\centering
S = \frac{1}{N} \sum_{i=1}^{N} \frac{b(i) - a(i)}{\max(a(i), b(i))}
\end{equation}

\noindent where $N$ is the total number of samples, $a(i)$ is the mean distance between sample $i$ and all other points in the same cluster, and $b(i)$ is the smallest mean distance between sample $i$ and any other points in any other cluster, representing the separation from neighboring clusters. Silhouette score has a range between -1 and 1. 

\subsubsection{Eigengap Heuristic}
Eigengap heuristic is a heuristic method for selecting the optimal number of clusters in 
clustering methods. According to the matrix perturbation theory \cite{stewart_matrix_1990}, if the eigengap of affinity matrix's graph Laplacian is larger, then the subspaces spanned by its corresponding eigenvectors will be closer to being ideal. Let $\lambda_{i}$ and $\lambda_{i+1}$ be two consecutive eigenvalues of the Laplacian matrix of the affinity matrix, their eigengap is:

\begin{equation}
\centering
\delta_{i} = |\lambda_{i+1} - \lambda_{i}|
\end{equation}

\noindent Let $N$ be the total number of samples in the dataset, ${\delta_{1}, ..., \delta_{N-1}}$ is then the set of all possible eigengap values, and the ideal number of clusters $K$ can be derived as follows: 

\begin{equation}
\centering
K = argmax(\delta_{i})
\end{equation}

\subsubsection{Davies-Bouldin Index}
Davies-Bouldin index is another quantitative measure of the clustering quality with similar intuition as the silhouette score of minimizing the within cluster distances and maximizing the between cluster distances. The Davies-Bouldin index can be written as the following formula:
\begin{equation}
\centering
DB = \frac{1}{N} \sum_{i=1}^{N} \max_{i \neq j} \frac{d(i) + d(j)}{D(c_i, c_j)}
\end{equation}

\noindent where DB is the Davies-Bouldin index of the clustering, $N$ is the number of clusters, $d(i)$ and $d(j)$ are the within-cluster distances between cluster $i$ and it's most similar cluster $j$, and $D(c_i, c_j)$ is the distance between the centroids of cluster $i$ and $j$. A lower DB score means better clustering quality.

\subsubsection{Calinski-Harabasz Index}
Calinski-Harabasz Index (also known as Variance Ratio Criterion) evaluates the clustering quality by estimating the ratio between "between cluster variance" and "within cluster variance". It can be described with the following formula:
\begin{equation}
\centering
CH(K) = \frac{\sum_{k=1}^{K} n_k \cdot D(c_k, c) / (K - 1)}{\sum_{k=1}^{K} \sum_{i=1}^{n_k} D(x_i, c_k) / (N - K)}
\end{equation}

\noindent where CH(K) is the Calinski-Harabasz index for cluster $K$, $n_k$ is the number of samples in cluster $K$, $D(c_k, c)$ is the distance between the centroid of cluster $K$ and the centroid of all samples, and $x_i$ is a sample in cluster $K$. A higher CH score indicates better clustering quality.

\subsubsection{Combining Different Model Selection Criteria}
We propose to combine the above four different model selection criteria by first computing a confidence score for each criterion on the motion affinity matrix for a range of possible number of motions that may be present in the scene, then selecting the number with the highest average confidence score as the number of motion groups present in the scene, and use this as the number of clusters to perform spectral clustering.

To calculate the above model selection metrics given a motion affinity matrix, we first need to transform the affinity matrix into a "distance matrix", due to the fact that the silhouette score, Davies-Bouldin index and Calinski-Harabasz index operate on distances among samples and clusters, instead of their similarities. Since all motion affinity matrices are normalized (i.e., having pairwise object motion affinity values between 0 and 1), we simply compute the pairwise object motion distance as $1 - affinity$. Then, we use this distance matrix to compute the normalized confidence score corresponding to each of the three criteria. Each normalized confidence score is valued between 0 and 1 with higher value indicating higher confidence. For eigengap heuristic, since it is not a quantitative measurement of the clustering quality, we compute its confidence score by checking how close the current number of motion clusters is to the optimal number of motion clusters (the one with the largest eigengap). Since we have a predefined range of how many motions may be present in the scene, it is easy to compute a normalized confidence score for eigengap heuristic in the same way as other criteria.

The above method is works for automatic model selection given a single motion affinity matrix. In cases of multiple multiple affinity matrices, we propose to first add these affinity matrices together, then perform row normalization \cite{von_luxburg_tutorial_2007} to obtain a normalized fused affinity matrix. We then perform the same procedure as above to infer the optimal number of motions using the fused affinity matrix.

\section{Experiments}
We evaluate our model selection method on the KT3DMoSeg dataset, which is a challenging monocular motion segmentation dataset focusing on real world scenes with strong motion degeneracy and motion parallax. The dataset contains manually corrected point trajectories obtained from an optical flow tracker on 22 video sequences selected from the KITTI dataset \cite{geiger_are_2012}. Each video sequence contains 2 to 5 different motion groups. Our evaluations are based on three criteria: 1) The mean squared error (MSE) of each method in predicting the number of motions; 2) The percentage of video sequences each method succeeds in predicting the exact number of motions correctly; 3) The overall motion segmentation error rates of different model selection techniques, versus that achieved by the baseline motion segmentation pipeline given the groundtruth number of motion clusters. The overall motion segmentation error rate is computed as the average error rate of all 22 sequences in the dataset, and the error rate of each sequence is computed as the percentage ratio between the number of wrongly clustered trajectories and the total number of trajectories in the sequence. This metric is adopted from \cite{xu_motion_2018}. 

The motion segmentation pipeline computes two motion affinity matrices using epipolar geometry and optical flow respectively. We evaluate our motion selection method both individually on each of the two matrices, and on the fused affinity matrix. The fused affinity matrix is computed by taking the element-wise mean of the two matrices.

We also compare our proposed method of combining different model selection criteria with a consensus voting method and random guessing. The consensus voting method chooses the most frequent optimal number of motion clusters computed by all four criteria. If there is not a most frequent number, it chooses the smaller median value. The random guessing method simply uses a random number between 2 and 5 (inclusive) as the number of motions for each video sequence.

\begin{table}[h]
    \centering
    \begin{minipage}{\linewidth}
    \centering
        \caption{MSE of different model selection methods on different motion affinity matrices (higher is better). Aff. F is the motion affinity matrix obtained using epipolar geometry, Aff. OC is the motion affinity matrix obtained using optical flow, and Fused Aff. is the fused motion affinity matrix by taking the mean of the affinity scores of these two matrices}
        \label{tab: Model Selection MSE}
        \renewcommand{\arraystretch}{1.3}
        \begin{tabularx}{\textwidth}{l>{\centering\arraybackslash}X>{\centering\arraybackslash}X>{\centering\arraybackslash}c>{\centering\arraybackslash}X}
            \toprule
            \textbf{Methods} & \textbf{Aff. F} & \textbf{Aff. OC} & \textbf{Fused Aff.} & \textbf{Avg. MSE} \\
            \midrule
            Silhouette  & 1.364         & \textbf{1.136}    & \underline{1.091} & \underline{1.197} \\
            Eigengap    & 1.318         & 1.455             & 1.636             & 1.470 \\
            DB          & \textbf{1.091}& 1.818             & 1.500             & 1.470 \\
            CH          & 1.364         & \underline{1.318} & 1.227             & 1.303 \\
            Random      & 3.909 & 2.455             & 3.091    & 3.152 \\
            Voting      & \textbf{1.091}& 1.455             & \textbf{1.046}    & \underline{1.197} \\
            Average     & \textbf{1.091}& 1.364             & \underline{1.091} & \textbf{1.182} \\
            \bottomrule
        \end{tabularx}
        \label{tab: Quantitative Results}
    \end{minipage}
\end{table}

\begin{table}[h]
    \centering
    \begin{minipage}{\linewidth}
    \centering
        \caption{Prediction accuracy of different model selection methods on different motion affinity matrices (higher is better).} 
        \label{tab: Model Selection Accuracy}
        \renewcommand{\arraystretch}{1.2}
        \begin{tabularx}{\textwidth}{l>{\centering\arraybackslash}X>{\centering\arraybackslash}X>{\centering\arraybackslash}c>{\centering\arraybackslash}X}
            \toprule
            \textbf{Methods} & \textbf{Aff. F} & \textbf{Aff. OC} & \textbf{Fused Aff.} & \textbf{Avg. Acc.} \\
            \midrule
            Silhouette  & \textbf{54.55}& \underline{54.55} & 59.09             & \textbf{56.06} \\
            Eigengap    & 45.45         & \textbf{59.09}    & 40.91             & 48.48 \\
            DB          & 54.55         & 31.82             & 40.91             & 42.42 \\
            CH          & \textbf{54.55}& 31.82             & \textbf{68.18}    & 51.52 \\
            Random      & 31.82         & 31.82             & 27.27             & 30.30 \\
            Voting      & 54.55         & 40.91             & 59.09             & 51.52 \\
            Average     & \textbf{54.55}& 45.45             & \underline{63.64} & \underline{54.54} \\
            \bottomrule
        \end{tabularx}
        \label{tab: Quantitative Results}
    \end{minipage}
\end{table}

\begin{table}[H]
    \centering
    \begin{minipage}{\linewidth}
    \centering
        \caption{Overall motion segmentation error rates of different model selection methods vs. the error rate obtained from known groundtruth number of motions (lower is better)}
        \label{tab: Motion Segmentation Error Rate}
        \renewcommand{\arraystretch}{1.2} 
        \begin{tabularx}{\textwidth}{l>{\centering\arraybackslash}X>{\centering\arraybackslash}X>{\centering\arraybackslash}c>{\centering\arraybackslash}c}
            \toprule
            \textbf{Methods} & \textbf{Aff. F} & \textbf{Aff. OC} & \textbf{Fused Aff.} & \textbf{Avg. Error} \\
            \midrule
            Silhouette  & 15.99             & \textbf{19.68}    & 12.78             & \underline{16.16} \\ 
            Eigengap    & 16.36             & 25.01             & 16.47             & 19.28 \\ 
            DB          & \underline{14.70} & 26.16             & 14.11             & 18.32 \\
            CH          & 18.03             & 26.88             & 12.09             & 19.01 \\ 
            Random      & 27.05             & 26.08             & 21.54             & 24.89 \\
            Voting      & 15.06             & 24.01             & \underline{12.04} & 17.04 \\
            Average     & \textbf{13.89}    & \underline{20.59} & \textbf{12.03}    & \textbf{15.50} \\
            \midrule
            Baseline & 9.86 & 13.47 & 5.78 & 9.71 \\
            \bottomrule
        \end{tabularx}
        \label{tab: Quantitative Results}
    \end{minipage}
\end{table}

Table \ref{tab:  Model Selection MSE} shows the mean squared errors of different model selection methods on different motion affinity matrices. Our proposed method (Average) achieves the best overall result in predicting the number of motions using the fused affinity matrix, followed by the consensus voting method and the silhouette method. 

Table \ref{tab:  Model Selection Accuracy} shows the accuracy of predicting the exact number of motions from different model selection methods on different motion affinity matrices. Silhouette score achieves the best result in terms of correctly predicting the exact number of motions in the scene. Our proposed method (Average) achieves the second best result.

Table \ref{tab:  Motion Segmentation Error Rate} shows the final motion segmentation error rate of the motion segmentation pipeline using different model selection methods. Our proposed method achieves the best results on two out of three types of motion affinity matrices, close to the baseline which takes the groundtruth number of motions as input. The silhouette method and the consensus voting method are the second and third best methods, indicating their strengths as well, which is consistent with the results in Table \ref{tab:  Model Selection MSE} and \ref{tab:  Model Selection Accuracy}.

To further investigate the strengths and weaknesses of our method, we also analyze the evaluation results in more detail by comparing the performance of each method on sequences containing different numbers of motions. Out of the 22 sequences, 12 sequences contain 2 motion groups, 4 sequences contain 3 motion groups, 5 sequence contains 4 motion groups and 1 sequence contains 5 motion groups. We show the MSE and the overall motion segmentation error rate of each method on sequences containing each number of motions in table \ref{tab:  Model Selection MSE by number of motions} and table \ref{tab: Model Selection Error Rates by number of motions} respectively. The results are evaluated using only the fused affinity matrix since the best motion segmentation results are usually obtained by fusing both affinity matrices together, thereby making the fused matrix more important and useful. 

Our proposed method performs well when there are only 2 motion groups in the sequence, which accounts for around half of the dataset. For sequences containing 3 or 5 motion groups, our method also performs decently well, being above average. However, for sequences containing 4 motion groups, our method does not perform well. In fact, most methods do not perform well on these sequences. This is mostly likely due to the fact that these video sequences generally contain more challenging scenes (e.g., more motion degeneracy or motion parallax) for the motion segmentation algorithm, resulting in motion affinity matrices of lower quality. As shown in table \ref{tab: Model Selection Error Rates by number of motions}, the baseline method where the groundtruth number of motions is given also performs worst on these sequences. 

\begin{table}[h]
    \centering
    \begin{minipage}{\linewidth}
    \centering
        \caption{MSE of different model selection methods on different numbers of motions. Avg. MSEs are computed using all 6 methods. Evaluated on the fused motion affinity matrix only.}
        \label{tab: Model Selection MSE by number of motions}
        \renewcommand{\arraystretch}{1.2}
        \begin{tabularx}{0.9\textwidth}{l|>{\centering\arraybackslash}X>{\centering\arraybackslash}X>{\centering\arraybackslash}X>{\centering\arraybackslash}X}
            \toprule
            \multirow{2}{*}{\makecell{\textbf{Methods}}} &
            \multicolumn{4}{c}{\textbf{Number of Motions}} \\
            \textbf{} & 
            \textbf{2} & \textbf{3} & \textbf{4} & \textbf{5} \\
           
            \midrule
            Silhouette  & 0.00  & 1.75  & 3.20 & 1.00 \\
            Eigengap    & 0.33  & 0.75  & 4.0  & 9.00 \\
            DB          & 1.167 & 3.25  & 1.00 & 1.00 \\
            CH          & 0.75  & 1.50  & 2.40 & 0.00 \\
            Voting      & 0.00  & 1.50  & 3.20 & 1.00 \\
            Average     & 0.00  & 1.75  & 3.20 & 1.00 \\
            \midrule
            Avg. MSE    & 0.375  & 1.75  & 2.83 & 2.17 \\
            \bottomrule
        \end{tabularx}
        \label{tab: Quantitative Results}
    \end{minipage}
\end{table}

\begin{table}[h]
    \centering
    \begin{minipage}{\linewidth}
    \centering
        \caption{Overall error rates of different model selection methods on different numbers of motions. Avg. Errors are computed using all 6 methods. Evaluated on the fused motion affinity matrix only.}
        \label{tab: Model Selection Error Rates by number of motions}
        \renewcommand{\arraystretch}{1.2}
        \begin{tabularx}{0.9\textwidth}{l|>{\centering\arraybackslash}X>{\centering\arraybackslash}X>{\centering\arraybackslash}X>{\centering\arraybackslash}X}
            \toprule
            \multirow{2}{*}{\makecell{\textbf{Methods}}} &
            \multicolumn{4}{c}{\textbf{Number of Motions}} \\
            \textbf{} & 
            \textbf{2} & \textbf{3} & \textbf{4} & \textbf{5} \\
            \midrule
            Silhouette  & 6.10   & 20.03  & 23.43 & 10.52 \\
            Eigengap    & 10.74  & 24.09  & 22.51  & 24.61 \\
            DB          & 10.40 & 20.44  & 18.67 & 10.52 \\
            CH          & 7.95  & 17.72  & 17.75 & 10.96 \\
            Voting      & 6.10  & 18.21  & 21.67 & 10.52 \\
            Average     & 6.10  & 18.16  & 21.67 & 10.52 \\
            \midrule
            Avg. Error  & 7.90  & 19.78  & 20.95 & 12.94 \\
            \midrule
            Baseline    & 3.31  & 8.23  & 13.75 & 6.04 \\
            \bottomrule
        \end{tabularx}
        \label{tab: Quantitative Results}
    \end{minipage}
\end{table}
\vspace{-1mm}

\section{Conclusion}
We proposed a unified model selection technique for spectral clustering based motion segmentation methods, to automatically infer the number of motions in the scene. We combine four existing model selection criteria by computing custom confidence scores on a range of possible numbers of motions, and select the number with the highest average confidence among all four criteria as the optimal number of motions. This inferred number is then used to perform spectral clustering to obtain the final motion segmentation. Our method was tested with a state-of-the-art motion segmentation method on the challenging KT3DMoSeg dataset and achieved competitive results, achieving an overall error rate close to the baseline which takes the groundtruth number of motions as input.

\clearpage
\bibliographystyle{IEEEtran}  
\bibliography{references.bib}
\end{document}